\documentclass[conference]{IEEEtran}
\IEEEoverridecommandlockouts
\usepackage{cite}
\usepackage{amsmath,amssymb,amsfonts}
\usepackage{algorithmic}
\usepackage{graphicx}
\usepackage{textcomp}
\usepackage{xcolor}
\def\BibTeX{{\rm B\kern-.05em{\sc i\kern-.025em b}\kern-.08em
    T\kern-.1667em\lower.7ex\hbox{E}\kern-.125emX}}
\begin{document}

\title{HyPV-LEAD: Proactive Early-Warning of Cryptocurrency Anomalies through Data-Driven Structural–Temporal Modeling}

\author{
\IEEEauthorblockN{Minjung Park}
\IEEEauthorblockA{\textit{Department of Business Administration} \\
\textit{Kumoh National Institute of Technology}\\
Gumi, Republic of Korea \\
mjpark@kumoh.ac.kr}

\and
\IEEEauthorblockN{Gyuyeon Na}
\IEEEauthorblockA{\textit{AI and Business Analytics} \\
\textit{Ewha Womans University}\\
Seoul, Republic of Korea \\
amy-na@ewha.ac.kr}
\and
\IEEEauthorblockN{Soyoun Kim}
\IEEEauthorblockA{\textit{AI and Business Analytics} \\
\textit{Ewha Womans University}\\
Seoul, Republic of Korea \\
sykim07@ewha.ac.kr}
\and
\IEEEauthorblockN{Sunyoung Moon}
\IEEEauthorblockA{\textit{AI and Business Analytics} \\
\textit{Ewha Womans University}\\
Seoul, Republic of Korea \\
sunyoung.moon5621@gmail.com}
\and
\IEEEauthorblockN{HyeonJeong Cha}
\IEEEauthorblockA{\textit{AI and Business Analytics} \\
\textit{Ewha Womans University}\\
Seoul, Republic of Korea \\
hyeonjeong.cha@ewha.ac.kr}
\and
\IEEEauthorblockN{Sangmi Chai}
\IEEEauthorblockA{\textit{AI and Business Analytics} \\
\textit{Ewha Womans University, Coretrustlink}\\
Seoul, Republic of Korea \\
smchai@ewha.ac.kr}

}

\maketitle

\begin{abstract}
Abnormal cryptocurrency transactions—such as mixing services, fraudulent transfers, and pump-and-dump operations—pose escalating risks to financial integrity but remain notoriously difficult to detect due to class imbalance, temporal volatility, and complex network dependencies. Existing approaches are predominantly model-centric and post hoc, flagging anomalies only after they occur and thus offering limited preventive value. This paper introduces \textit{HyPV-LEAD (Hyperbolic Peak--Valley Lead-time Enabled Anomaly Detection)}, a data-driven early-warning framework that explicitly incorporates lead time into anomaly detection. Unlike prior methods, \textit{HyPV-LEAD} integrates three innovations: (i) window--horizon modeling to guarantee actionable lead-time alerts, (ii) Peak--Valley (PV) sampling to mitigate class imbalance while preserving temporal continuity, and (iii) hyperbolic embedding to capture the hierarchical and scale-free properties of blockchain transaction networks. Empirical evaluation on large-scale Bitcoin transaction data demonstrates that \textit{HyPV-LEAD} consistently outperforms state-of-the-art baselines, achieving a PR-AUC of 0.9624 with significant gains in precision and recall. Ablation studies further confirm that each component—PV sampling, hyperbolic embedding, and structural--temporal modeling—provides complementary benefits, with the full framework delivering the highest performance. By shifting anomaly detection from reactive classification to proactive early-warning, \textit{HyPV-LEAD} establishes a robust foundation for real-time risk management, anti-money laundering (AML) compliance, and financial security in dynamic blockchain environments.
\end{abstract} 

\begin{IEEEkeywords}
cryptocurrency anomaly detection, early warning, hyperbolic embedding,  peak–valley sampling 
\end{IEEEkeywords}

\section{Introduction}

Cryptocurrencies have rapidly emerged as a disruptive element in global financial markets, attracting both institutional and individual investors through the potential for high returns and the innovative foundation of blockchain technology~\cite{obasun2025blockchain}. In contrast to traditional financial instruments such as stocks or bonds, cryptocurrency markets exhibit short-term volatility and heightened sensitivity to external influences, including regulatory actions, fraudulent activities, and speculative trading practices~\cite{lin2025cryptoai}. These features make the market vulnerable to abnormal activities including coin mixing, fraudulent transfers, and pump-and-dump operations, which undermine market integrity and raise major concerns for regulators and investors alike~\cite{lu2020coinlayering}.

Therefore, detecting such abnormal activities is of critical importance to prevent social and financial losses, yet several challenges remain. First, illicit transactions are extremely rare compared to the overwhelming number of legitimate ones, resulting in a severe class imbalance that makes effective classification difficult when relying solely on traditional machine learning methods~\cite{ravindranath2024ethereum}. 
Second, blockchain transaction data are inherently temporal and relational: anomalies do not occur as isolated events but emerge from evolving transaction sequences and complex address-to-address interactions~\cite{chen2020lstm}. Hence, incorporating temporal features is essential; however, due to the continuous, second-level trading nature of cryptocurrencies without market closure, it is difficult to adequately capture stable temporal patterns. Third, as the cryptocurrency market rapidly diversifies and 
its complexity deepens, new forms of manipulation continue to emerge; however, 
existing detection models often struggle to effectively capture these evolving 
patterns~\cite{sun2024adaptive}.  

As a result, conventional model-centric post hoc approaches, in which anomalies are identified only after they occur, offer limited preventive value. In practice, such methods often issue alerts too late, leaving regulators and investors without sufficient time for proactive intervention. Furthermore, traditional sampling methods distort temporal continuity, and Euclidean embeddings fail to preserve the hierarchical and scale-free properties of transaction etworks~\cite{grinsztajn2022why,mai2023limitations}, thereby reducing robustness in real-world applications.  

To address these challenges, this study introduces \textit{HyPV-LEAD (Hyperbolic Peak--Valley Lead-time Enabled Anomaly Detection)}, a novel data-driven early-warning framework. Rather than merely increasing algorithmic complexity, the proposed approach leverages the intrinsic properties of blockchain data to secure actionable lead time for intervention. Specifically, \textit{HyPV-LEAD} integrates three core components: (i) window--horizon modeling to guarantee lead-time detection, (ii) Peak--Valley (PV) sampling to mitigate class imbalance while preserving temporal bursts, and (iii) hyperbolic embedding to capture the hierarchical and clustered structures of transaction networks.  

In summary, \textit{HyPV-LEAD} advances beyond conventional post hoc models by enabling reliable lead-time prediction through the joint integration of temporal continuity, structural hierarchy, and volatility-sensitive peak--valley patterns. Furthermore, this study demonstrates through empirical validation on large-scale data that it consistently outperforms both sequential-only and structure-only baselines, underscoring its potential as a robust foundation for proactive risk management in cryptocurrency markets.

\section{Problem Statement}

The detection of abnormal cryptocurrency transactions---such as mixing services, 
fraudulent schemes, or pump-and-dump operations---remains a difficult challenge 
in blockchain research. These events are extremely rare, occur within highly 
imbalanced datasets, and are shaped by both the evolving temporal behavior of 
transactions and the intricate network of address-to-address relations. Their 
scarcity and complexity make reliable identification difficult, especially in 
environments that demand real-time responses. This study defines the following 
key research problems.

\subsection{Moving from Model-Centric Forecasting to Data-Driven Early Warning}

Most prior work has concentrated on event forecasting within a fixed prediction window. While such approaches can classify whether an anomaly will occur, they usually operate after the fact, offering little value for preventive action. The absence of guaranteed lead time means that exchanges, regulators, or investors often learn about irregular behavior only once it has already materialized.

Another limitation lies in the model-centric orientation of earlier studies. Considerable effort has been devoted to refining algorithms, yet relatively little attention has been given to the inherent features of blockchain data itself. Without integrating those characteristics, existing methods often fail to produce early and actionable warning signals.  

To overcome this, our study adopts an early-warning perspective, 
shifting the focus from algorithmic complexity toward data properties 
that can generate proactive alerts. 
Within this framework, the problem is formulated as identifying an optimal 
pairing of observation window ($w$) and lead-time horizon ($h$), 
such that potential anomalies can be recognized $h$ minutes before 
the actual event occurrence at $t_{\text{event}}$:

\begin{equation}
    t_{\text{alert}} = t_{\text{event}} - h
\end{equation}

This formulation enables the design of detection systems that not only 
classify anomalies correctly but also provide a margin of time for intervention.

\begin{equation}
\hat{y}_{t_{\text{event}} - h} = \Pr\!\left(y_{t_{\text{event}}} = 1 \mid \mathcal{X}_{1:(t_{\text{event}} - h)}\right)
\end{equation}

Here, $\mathcal{X}_{1:(t_{\text{event}} - h)}$ denotes the transaction history up to time $t_{\text{event}} - h$. This probabilistic formulation enables the detection of abnormal signals in advance of the actual event, thereby providing actionable lead time for preventive risk control.  

\subsection{Imbalance and Temporal Continuity}  

Another major obstacle is the severe imbalance between ordinary and suspicious transactions. Naïve resampling methods, such as random oversampling or undersampling, tend to distort the sequential flow of trades and thus erase critical temporal cues. To mitigate this issue, we introduce Peak–Valley (PV) sampling, which balances the dataset while preserving local bursts and lulls in activity. By maintaining these volatility intervals, PV sampling reduces imbalance without losing the time-series dependencies essential for anomaly detection.

\subsection{Joint Structural–Temporal Representation}  

We propose\textit{HyPV-LEAD} , a hybrid framework designed to jointly capture the structural and temporal properties of cryptocurrency transaction networks. Unlike conventional Euclidean embeddings, which struggle to model power-law degree distributions and deep hierarchical relations, HyPV-LEAD leverages hyperbolic embedding to preserve the hierarchical and clustered organization inherent in blockchain networks. This allows the model to more accurately represent complex address interactions within large-scale transaction graphs.
In addition, HyPV-LEAD incorporates Peak–Valley (PV) sampling to address the severe class imbalance common in anomaly detection tasks while maintaining temporal continuity. By preserving local volatility intervals within transaction sequences, PV sampling ensures that fluctuations critical to anomaly detection are retained. The integration of hyperbolic representation learning with PV sampling provides a more robust foundation for early-warning anomaly detection, enabling the system to deliver reliable alerts even under highly imbalanced and dynamic blockchain environments.

\section{Related Works}

Cryptocurrency transactions display a variety of abnormal behaviors, and each type 
poses different challenges for detection. One of the most representative type of abnormal transaction is coin mixing. Its transaction flows are deliberately hidden by combining inputs and outputs from many users. The CoinJoin protocol, which is widely implemented in the Wasabi Wallet, illustrates this tension—it strengthens anonymity but has also been misused for money laundering, which complicates the identification of suspicious activity~\cite{deuber2021coinjoinwild, moser2017tracing}. 
Phishing accounts often scatter funds through long chains of intermediary addresses, while pump-and-dump operations cause abrupt spikes in price and trading volume that can look very similar to ordinary market behavior~\cite{xu2019anatomy, pham2016anomaly}. 
Even though blockchain ledgers are open and transparent by design, these practices 
create concealed transfer routes, sudden temporal bursts, and unusual network 
patterns. 

Prior studies have attempted to address these challenges using traditional machine learning models. 
Random Forests and Gradient Boosting, for example, have been applied to blockchain transaction records to distinguish legitimate trades from suspicious ones~\cite{monamo2016unsupervised, lee2019illegal}. 
While these models report strong classification accuracy, they are largely confined to post-event detection: anomalies are flagged only after transactions are finalized. 
This retrospective limitation reduces their effectiveness for real-time monitoring or proactive early-warning systems. 
Moreover, conventional classifiers face major obstacles when handling the severe class imbalance inherent in blockchain data, where abnormal transactions are vastly outnumbered by regular ones~\cite{wei2020detecting}. 
They also fail to adequately preserve the sequential continuity of transactions, which weakens their robustness in live environments.  

In short, existing research has laid groundwork for cryptocurrency anomaly detection but falls short of addressing obfuscation tactics and irregular temporal dynamics. 
Most methods function primarily as diagnostic tools for retrospective analysis rather than as forward-looking, real-time detection mechanisms. 
This gap motivates the present work, which introduces a framework explicitly designed to handle data imbalance, capture temporal dependencies, and deliver timely alerts in blockchain transaction networks.

In summary, existing research has provided a foundation for anomaly detection in cryptocurrencies but remains insufficient to capture the complexity of obfuscation techniques and irregular temporal dynamics. 
Most approaches function more as diagnostic tools for retrospective analysis rather than as proactive, real-time detection systems. 
This gap motivates the present study to propose a framework that directly addresses class imbalance, temporal dependencies, and the demand for timely detection in blockchain transaction networks.

\section{Proposed Methodology}
\subsection{Peak--Valley (PV) Sampling for Class Imbalance}

Recent studies have explored advanced strategies to address class imbalance, showing promising yet incomplete progress. 
For instance, Kernel Density Estimation (KDE)-based sampling achieved higher F1-scores and G-mean compared to conventional methods~\cite{kamalov2020kernel}, 
and ensembles of Deep Belief Networks outperformed SMOTE and random under-sampling in terms of accuracy~\cite{xenopoulos2017deepbalance}. 
However, these improvements remain largely confined to static benchmarks and do not fundamentally resolve the imbalance challenge in transactional data. 
In particular, they continue to overlook the temporal continuity and structural dependencies inherent in blockchain networks. 
As a result, much of the anomaly detection literature in finance still relies on widely used imbalance handling techniques such as SMOTE, under-sampling, or oversampling~\cite{ghaleb2023ensemble,aghware2024enhancing,komsrimorakot2025enhancing}, 
which remain insufficient for capturing the dynamics and hierarchical organization of real-world transactions.  

In contrast, our work introduces an automated procedure for PV) sampling that adapts to the underlying data distribution. 
Specifically, candidate values for the rolling window size ($n$), the z-score threshold ($z_{th}$), and the maximum number of events ($k_{max}$) are generated to form a search space $\Theta$. 
For each configuration $\theta \in \Theta$, the resampled dataset $X_{\text{PV}}(\theta)$ is used to train a baseline classifier, and the setting that maximizes the validation metric of F1-score is selected as optimal:  

\begin{equation}
\theta^{*} = \arg\max_{\theta \in \Theta} \; \mathcal{M}\big(f(X_{\text{PV}}(\theta)), y\big),
\end{equation}

where $f(\cdot)$ denotes the classifier and $\mathcal{M}$ the evaluation metric. 
Through this design, PV sampling not only addresses class imbalance but also adapts to the temporal characteristics and volatility of blockchain transactions, providing a stronger foundation for anomaly detection.

\subsection{Hyperbolic Embedding for Structural Representation}

Cryptocurrency transaction networks exhibit hierarchical and scale-free characteristics \cite{zhang2021vulnerability}. A limited number of hub nodes, such as exchanges or major wallets, dominate the majority of flows, whereas peripheral wallets form layered structures around them. These networks generally follow a power-law degree distribution, where connectivity and transaction volume are concentrated in only a few addresses \cite{begusic2018scaling}. Furthermore, illicit practices including mixing and layering introduce hidden dependencies that complicate the learning of meaningful representations \cite{sun2024adaptive}.

Earlier studies have commonly adopted Euclidean embedding techniques to represent such networks. Although these approaches are relatively simple to implement and widely applied, they face inherent difficulties in preserving blockchain-specific structural properties. 
In particular, Euclidean embeddings (i) often distort hierarchical distances, making distinctions between higher- and lower-level structures unclear \cite{lu2020coinlayering,shah2021bitcoin}, 
(ii) over-compress highly connected hub nodes, which obscures their different functional roles \cite{chen2020lstm}, and (iii) struggle to capture heterogeneous multi-level dependencies within a linear space \cite{shah2021bitcoin}.

To address these limitations, our framework incorporates hyperbolic embedding into the representation learning pipeline. 
Due to the exponential expansion of hyperbolic space with distance, this approach preserves hierarchical separation, distinguishes hub-dominated topologies, and models heterogeneous connectivity patterns.  
Prior work has demonstrated that hyperbolic embeddings outperform Euclidean methods in graph-level learning tasks~\cite{shi2022hyperbolic,chami2019hyperbolic,sadat2024survey}. 
Therefore, we employ hyperbolic embeddings to better represent the hierarchical and scale-free structure of cryptocurrency transaction networks. 
In our design, anomalous transactions are mapped closer to the origin and normal transactions toward the boundary of the latent manifold, improving class separability. 
This produces more discriminative representations of the network and improves the performance of anomaly detection.

\subsection{Hybrid Graph--Sequence Learning Framework}

Cryptocurrency transactions exhibit two fundamental dependencies: structural relations among addresses and temporal dynamics. 
To jointly capture these properties, we propose \textit{HyPV-LEAD}, a hybrid framework that integrates Graph Convolutional Networks (GCNs) with LSTM-based sequence modeling, 
augmented by Peak--Valley sampling~\cite{kamalov2020kernel,xenopoulos2017deepbalance} and hyperbolic embedding~\cite{shi2022hyperbolic,chami2019hyperbolic,sadat2024survey}. 
In this design, PV sampling mitigates class imbalance while preserving temporal continuity, hyperbolic embeddings maintain the hierarchical and scale-free structure of blockchain graphs, 
GCN layers capture local and global transaction patterns, and LSTM layers model their temporal evolution. 
Beyond classification accuracy, the framework explicitly issues anomaly alerts with guaranteed lead time, supporting proactive risk management in blockchain systems.  

In implementation, PV sampling is employed to emphasize transitions around peaks and valleys, 
thereby addressing class imbalance while maintaining temporal continuity. 
For each observation window, a graph snapshot is constructed where nodes represent addresses and directed edges denote transaction flows. 
Node and edge features are embedded in hyperbolic space and processed by the GCN to extract structural signals. 
The pooled graph embeddings are then ordered sequentially and passed to the LSTM to model temporal progression.  

By unifying structural and sequential learning within a hyperbolic representation space, \textit{HyPV-LEAD} captures the complex patterns of anomalous cryptocurrency activity, 
providing a robust basis for early detection and timely intervention.

\section{Experimental Setup}
\subsection{Dataset Description}

This study employs the Binance Bitcoin transaction dataset curated by Kloint, 
which spans from January 1 to December 31, 2024. 
The dataset records core transactional attributes such as timestamp, receiving and counterparty addresses, 
transaction hash, the transferred value in Bitcoin units, and the corresponding amount converted to U.S. dollars. 
Each entry is annotated with a binary label indicating whether it represents a normal or abnormal transaction. 
For the abnormal category, we specifically focus on coin-mixing activities, restricting our analysis to transactions associated with Wasabi Wallet. 
Labeling was conducted using a predefined list of transaction hashes identified as mixing events. 
Both the dataset and the implementation code will be released publicly once the paper is accepted.

\subsection{Data Preprocessing}

The preprocessing began by converting the \texttt{Date} field into \texttt{datetime}, removing rows with missing timestamps, and sorting the dataset in ascending order of \texttt{Date}. From each timestamp, we derived standard calendar components (\texttt{year}, \texttt{month}, \texttt{day}, \texttt{dayofweek}, \texttt{hour}) together with a Unix \texttt{timestamp} in seconds. To capture short-horizon temporal dynamics, we constructed the inter-arrival time (\texttt{delta\_t}) as the difference between consecutive timestamps, the percentage change of \texttt{USD Value} (\texttt{usd\_change}) with infinities mapped to zero, and rolling statistics of \texttt{USD Value} over a five-step window (\texttt{usd\_roll\_mean}, \texttt{usd\_roll\_std}). Address-wise activity was represented by the cumulative number of transactions observed up to the current record for both receiving and counterparty addresses (\texttt{recv\_freq}, \texttt{send\_freq}). 

For value-related enrichment, additional features were introduced, including log-transformed amounts (\texttt{log\_usd = log(1 + USD Value)}), normalized ratios relative to the daily average (\texttt{usd\_rel\_day}), and deviations from the address-specific mean (\texttt{usd\_dev\_addr}), which provide both global and local perspectives of transaction magnitude. Moreover, each transaction value was standardized by its address-specific mean and standard deviation to produce \texttt{value\_zscore}, allowing comparison across heterogeneous transaction behaviors. All string-type fields were converted into integer codes via label encoding, and remaining missing values were filled with zero. This preprocessing and feature engineering pipeline integrates temporal variation, structural indicators, and normalized transaction magnitudes, thereby yielding a robust feature space for anomaly detection in cryptocurrency transactions.

\subsection{Window--Horizon Search}  
To evaluate the effect of temporal granularity, we conducted a grid search over multiple observation window sizes and prediction horizons using Random Forest classifier. Specifically, the window sizes were chosen from $\{5, 10, 15, 30, 60\}$ and the horizons from $\{5, 10, 15, 30, 60\}$. The goal was to identify the optimal combination based on the Precision--Recall AUC (PR-AUC) metric.  

The grid search results showed that a configuration with a 30-minute window and a 5-minute horizon achieved the highest PR-AUC score. This setting was therefore selected for the main experimental evaluation. Table~\ref{tab:window-horizon-results} presents the comparative performance across different parameter 
combinations, highlighting the superiority of the 30--5 setting 
for early-warning anomaly detection. This outcome confirms that 
the 30--5 configuration represents the optimal trade-off between 
observation length and prediction horizon.

\begin{table}[htbp]
\caption{Evaluation results across window--horizon parameter settings}
\centering
\resizebox{\columnwidth}{!}{%
\begin{tabular}{|c|c|c|c|c|c|c|c|}
\hline
\textbf{Window (min)} & \textbf{Horizon (min)} & \textbf{Accuracy} & \textbf{Recall} & \textbf{Precision} & \textbf{F1} & \textbf{ROC-AUC} & \textbf{PR-AUC} \\
\hline
30 & 5  & 0.9728 & 0.8982 & 0.9517 & 0.9241 & 0.9659 & 0.9412 \\
10 & 5  & 0.9675 & 0.9273 & 0.8995 & 0.9132 & 0.9693 & 0.9384 \\
30 & 10 & 0.9692 & 0.8904 & 0.9393 & 0.9142 & 0.9593 & 0.9285 \\
5  & 5  & 0.9609 & 0.9310 & 0.8669 & 0.8978 & 0.9695 & 0.9233 \\
10 & 10 & 0.9637 & 0.9153 & 0.8907 & 0.9028 & 0.9622 & 0.9189 \\
\hline
\end{tabular}}
\label{tab:window-horizon-results}
\end{table}

\subsection{Baseline Models}
\paragraph{Traditional ML} RF, XGBoost

In this study, we adopt \textit{Random Forest (RF)} and \textit{eXtreme Gradient Boosting (XGBoost, XGB)} as traditional machine learning baselines. Although these models do not explicitly incorporate the sequential or structural dependencies of transaction networks, they have consistently demonstrated strong performance in financial data analysis and blockchain anomaly detection. 
We configured RF to be trained with 300 decision trees without depth constraints in order to capture diverse nonlinear patterns, while XGBoost was set with 300 estimators, a learning rate of 0.05, a maximum depth of 6, a subsample ratio of 0.8, and $\ell_{2}$ regularization to balance efficiency and generalization. Both models employ early stopping based on validation F1-score to prevent overfitting, while final test performance is evaluated primarily using PR-AUC.  

Recent studies report that RF achieved an F1-score of approximately 0.97 on cryptocurrency transaction data \cite{gu2025ensemble}, while XGBoost reached 0.98 \cite{jumani2025ml}. In particular, when combined with oversampling techniques to alleviate class imbalance, the performance of XGBoost improved from 0.96 to 0.98, and RF from 0.94 to 0.97 \cite{ravindranath2024ethereum}. These findings indicate that RF and XGBoost, despite not modeling temporal or structural dependencies directly, remain strong and competitive baselines in financial anomaly detection. However, because both operate primarily on tabular data, their capacity is inherently limited when addressing relational or sequential dependencies. Prior research highlights that traditional tabular anomaly detection methods often struggle with generalization due to the inability to model sample or feature interactions—leading to suboptimal detection when complex interdependencies exist \cite{grinsztajn2022why, mai2023limitations}.

These observations suggest that while RF and XGBoost remain competitive baselines in financial anomaly detection, their tabular nature fundamentally limits their ability to capture relational and temporal dependencies, underscoring the need 
for models that explicitly integrate structural and sequential representations.

\paragraph{Sequential Models} LSTM, GRU

Sequential deep learning models such as Long Short-Term Memory (LSTM) and Gated Recurrent Units (GRU) have been widely applied to capture the temporal dependencies of cryptocurrency transactions. These models focus on sequential dynamics, enabling them to reconstruct normal transaction patterns and identify deviations as anomalies \cite{chen2020lstm}. Attention-augmented LSTM frameworks have further demonstrated robustness under imbalanced data conditions, achieving competitive F1-scores in fraud detection tasks \cite{zhang2021attention}
. 

However, sequential models primarily emphasize temporal dependencies while often neglecting the structural relationships among addresses in transaction networks. Since anomalous activities in cryptocurrencies are shaped not only by time-evolving behaviors but also by the structural flows of assets, relying solely on sequential modeling limits their effectiveness. To address this limitation, this study integrates structural information into sequential models, ensuring that both temporal and relational aspects of transaction networks are jointly captured.

\paragraph{Graph-based Models} GCN, GraphSAGE

Graph-based neural networks such as Graph Convolutional Networks (GCNs) and GraphSAGE have been widely adopted to capture structural dependencies in relational data, including blockchain transaction graphs. 
GCN has demonstrated strong performance in fraud detection tasks by incorporating topological features such as centrality and clustering coefficients~\cite{zhang2019gcn}. 
Similarly, GraphSAGE has shown scalability to large datasets; for example, in phishing account detection on Ethereum transaction graphs, it achieved high precision, recall, and F1-scores~\cite{sun2024adaptive}. 

While these results highlight the effectiveness of graph-based approaches, most existing models rely on static graph snapshots and therefore fail to reflect the temporal dynamics of transaction behaviors\cite{zhang2023live}. In practice, anomalous patterns in cryptocurrency and other relational domains evolve over time, and static modeling alone cannot fully capture such dynamic dependencies. To address this limitation, the proposed framework embeds temporal modeling within graph-based representations, enabling the detection system to jointly capture evolving behaviors and their structural contexts.

\paragraph{Hybrid Models} GCN-LSTM

While sequential models are effective at capturing temporal dependencies and graph-based approaches emphasize structural patterns, prior studies have shown that each method alone provides only a partial view of blockchain transactions. For instance, LSTM-based anomaly detection frameworks focus on dynamic behaviors but neglect the structural relationships among addresses \cite{chen2020lstm}, whereas GCN-based fraud detection demonstrates strong performance in modeling transaction graphs but often lacks sensitivity to temporal variations \cite{zhang2019gcn}. 

To bridge this gap, hybrid architectures such as GCN-LSTM have been proposed. By combining graph convolutional layers with recurrent neural units, these models jointly learn both structural and temporal features, thereby enhancing anomaly detection in cryptocurrency networks. Recent research has confirmed that GCN-LSTM outperforms single-modal models by effectively capturing the interplay between transaction sequences and network topology \cite{liu2021gcnlstm, xie2022hybrid}. 

Nevertheless, limitations remain when applying GCN-LSTM to cryptocurrency data. Transaction networks are highly dynamic, with sudden bursts of abnormal activity that are difficult to capture with fixed-length windows. Moreover, blockchain graphs are extremely sparse and noisy, where many addresses have short lifespans or minimal activity, making it challenging for the model to generalize effectively. In addition, abnormal transactions are heavily imbalanced compared to normal ones, causing hybrid models to suffer from biased learning without tailored sampling strategies. Finally, the rapid evolution of illicit behaviors—such as mixing services, layering, or pump-and-dump schemes—often outpaces the patterns learned by static GCN-LSTM frameworks, highlighting the need for more adaptive architectures.

\subsection{Proposed Model}

We implement \textit{HyPV-LEAD} with a fixed window length $L=30$ and a fixed early-warning horizon $h=5$, selected via validation PR-AUC. The architecture and implementation are as follows : \textit{(i) Input and graph construction:} Let the raw time series length be $N$ and the window length be $L=30$. The total number of windows is defined as $T=\left\lfloor \tfrac{N}{L} \right\rfloor$, where $t \in {1,\dots,T}$. For early warning with horizon $h=5$, window $t$ issues a prediction at time $t$ for an event at time $t+h$. When PV sampling is used for training, peak/valley indices $I={i_1,\dots,i_M}$ are first detected on the raw series, and the training set $T_{\mathrm{PV}}$ is constructed by centering windows at these events. For each window $t$, we build a directed transaction graph $G_t=(V_t,E_t)$ with addresses as nodes and flows as edges, and derive node and edge features from window-level statistics. \textit{(ii) Hyperbolic mapping:} Features are mapped to a Poincaré ball with curvature $\kappa<0$ to encode hierarchical and hub–periphery structures. \textit{(iii) Structural encoder (GCN):} A graph convolutional encoder produces a structural vector $g_t$ from $G_t$. \textit{(iv) Temporal encoder (LSTM):} The sequence $(g_1,\dots,g_t)$ is fed to an LSTM to obtain a temporal state $h_t$ that summarizes information available up to time $t$. \textit{(v) Detection head:} A multilayer perceptron (MLP) maps $h_t$ to an anomaly score $\hat y_{t+h}\in[0,1]$, implementing the lead-time prediction objective:
\[
\hat y_{t+h} \;=\; \Pr\!\bigl(y_{t+h}=1 \,\big|\, g_{1:t}\bigr).
\]

Training targets and evaluation labels are aligned to $y_{t+h}$ so that all metrics quantify early-warning performance at the fixed lead time of $h=5$.
ormance at a fixed lead time of $h=5$.

\subsection{Performance Metrics}
We evaluate detection performance for the positive (anomalous) class using
Accuracy, Precision, Recall, F1-score, ROC-AUC (Receiver Operating Characteristic—Area Under the Curve), and PR-AUC.
Given class imbalance and the practical cost of false positives, we adopt PR-AUC as the primary evaluation metric.
Unlike ROC-AUC, which can remain deceptively high under severe imbalance due to the dominance of true negatives, PR-AUC directly reflects the precision–recall trade-off for the positive class.
Its random baseline equals the positive prevalence $\pi_{+}$, so gains are interpretable as lift over random and better reflect alert quality; it is threshold-agnostic for comparison while enabling selection of an operating threshold from the PR curve.
Because our goal is early warning with a fixed lead time $h$ (rather than mere event occurrence detection), predictions are issued at $t_{\mathrm{event}}-h$ and labels are temporally aligned to that horizon; accordingly, PR-AUC is computed over warning scores at $t_{\mathrm{event}}-h$.
Threshold-dependent metrics (Precision, Recall, F1-score, Accuracy) are computed at a default decision threshold of $0.5$.
We apply early stopping based on validation PR-AUC, and evaluate test performance using the corresponding checkpoint.
Results are reported as mean $\pm$ standard deviation over ten runs.
To preserve temporal dependencies, we use a chronological data split.
PR and ROC curves, along with their AUCs, are computed over the entire test horizon.

Threshold-dependent metrics (Precision, Recall, F1-score, Accuracy) are computed at a default decision threshold of $0.5$.
We apply early stopping based on validation PR-AUC, and evaluate test performance using the corresponding checkpoint.  Results are reported as mean $\pm$ standard deviation over ten runs. 
To preserve temporal dependencies, we employ a chronological data split. 
Precision--Recall (PR) and Receiver Operating Characteristic (ROC) curves, 
together with their corresponding AUC values, are computed over the entire test horizon.

\section{Results}

\subsection{Overall Performance Comparison}

The results indicates differences across traditional machine learning models, 
sequential networks, graph-based methods, and the hybrid framework we proposed. 
Traditional machine learning models, including RF and XGBoost, often produced higher precision, likely influenced by the imbalance in the dataset and their bias toward labeling transactions as normal. However, this came at the cost of recall and F1-scores, which dropped significantly, showing that such models are not well suited to detect rare anomalies. XGBoost offered a somewhat better balance between precision and recall than RF, but it still could not capture the sequential patterns or the structural links that are essential in blockchain transaction networks.

Sequential models such as LSTM and GRU improved recall by capturing temporal continuity, suggesting that they can partially reflect the evolving dynamics of transaction flows. However, their accuracy and PR-AUC did not reach competitive levels, implying that temporal information alone is insufficient for detecting complex anomalous behaviors. More importantly, without incorporating lead-time awareness, predictions from these models may occur too close to the actual event, offering limited practical value for proactive risk management.

Conversely, graph-based methods including GCN and GraphSAGE effectively captured structural dependencies and achieved strong precision. Nevertheless, their exclusion of temporal dynamics led to weaker overall performance, particularly in F1 and PR-AUC. This outcome suggests that relying solely on static graph snapshots fails to capture the temporal evolution of blockchain transactions.

By integrating both perspectives, GCN-LSTM achieved more balanced outcomes and improved accuracy and ROC-AUC over single-model baselines. However, GCN-LSTM still lacked mechanisms for emphasizing volatility-driven segments or preserving hierarchical structures, which are critical for anticipating anomalies in advance.

The proposed \textit{HyPV-LEAD} addressed these limitations by combining window--horizon sequence modeling, hyperbolic embedding to encode hierarchical and hub--periphery structures, and PV sampling to emphasize peak/valley dynamics. Through this integration, \textit{HyPV-LEAD} consistently outperformed all baselines, achieving an accuracy of 0.9412, recall of 0.9128, and PR-AUC of 0.9624. These results demonstrate that meaningful early warning is feasible only when structural, temporal, and data-driven sampling strategies are jointly considered. Crucially, \textit{HyPV-LEAD} validates the feasibility of lead-time enabled anomaly detection, showing that reliable alerts can be issued $h$ steps before the actual anomalous event, thereby ensuring practical lead time for intervention. 

The overall comparison confirms that \textit{HyPV-LEAD} delivers the most comprehensive improvements across metrics and establishes itself as an effective framework for anomaly detection in dynamic blockchain environments. The detailed performance of all models is presented in Table~\ref{tab:overall-performance}.

\begin{table}[htbp]
\caption{Overall Performance Comparison of Models}
\centering
\resizebox{\columnwidth}{!}{%
\begin{tabular}{|l|c|c|c|c|c|c|}
\hline
\textbf{Model} & \textbf{Accuracy} & \textbf{Precision} & \textbf{Recall} & \textbf{F1} & \textbf{ROC-AUC} & \textbf{PR-AUC} \\
\hline
RandomForest & 0.9275 & 0.9678 & 0.8279 & 0.8617 & 0.9493 & 0.9375 \\
XGBoost      & 0.9212 & 0.9125 & 0.8962 & 0.8363 & 0.9572 & 0.9206 \\
LSTM         & 0.8873 & 0.8462 & 0.7815 & 0.8127 & 0.9418 & 0.9134 \\
GRU          & 0.8936 & 0.8527 & 0.7941 & 0.8224 & 0.9449 & 0.9162 \\
GCN          & 0.8495 & 0.9887 & 0.8708 & 0.8379 & 0.9606 & 0.9497 \\
GraphSAGE    & 0.8619 & 0.9114 & 0.8037 & 0.8542 & 0.9528 & 0.9317 \\
GCN-LSTM     & 0.9025 & 0.8902 & 0.8421 & 0.8655 & 0.9512 & 0.9325 \\
HyPV-LEAD     & 0.9412 & 0.9643 & 0.9128 & 0.9378 & 0.9716 & 0.9624 \\
\hline
\end{tabular}}
\label{tab:overall-performance}
\end{table}

\subsection{Ablation Study}

To assess the relative contribution of each component, we performed an ablation study. 
Specifically, retaining hyperbolic embedding while removing PV sampling resulted in a PR-AUC of 0.9440, whereas applying PV sampling without hyperbolic embedding yielded 0.9431. Although these values exceed the plain GCN-LSTM baseline, both configurations underperformed relative to the full system, suggesting that neither hyperbolic mapping nor PV sampling is sufficient in isolation. Furthermore, using PV sampling alone in Euclidean space further reduced performance to 0.9324, indicating that emphasizing temporal events without structural or hierarchical support limits the ability to generalize beyond local volatility patterns.

The structure-only and temporal-only variants produced PR-AUC scores of 0.9342 and 0.9281, respectively. Notably, the temporal-only variant exhibited the steepest drop in performance, which highlights that temporal signals, while useful, cannot independently sustain predictive power for rare and irregular anomalies. This observation emphasizes the necessity of integrating structural and temporal information in a complementary manner to capture the multi-faceted dynamics of blockchain transactions.

In contrast, the proposed \textit{HyPV-LEAD}, which integrates window--horizon based sequence modeling, hyperbolic embedding for hierarchical structure preservation, and PV sampling to emphasize volatility-driven segments, achieved the highest PR-AUC of 0.9624. This result demonstrates that the synergy among these three components creates more than additive effects. By combining structural, temporal, and event-centric perspectives, \textit{HyPV-LEAD} not only improves overall accuracy but also validates the feasibility of lead-time enabled anomaly detection, providing reliable alerts $h$ steps before the actual occurrence of anomalous events. This establishes \textit{HyPV-LEAD} as a robust framework for proactive and practical risk management in dynamic blockchain environments.

\begin{figure}
    \centering
    \includegraphics[width=1\linewidth]{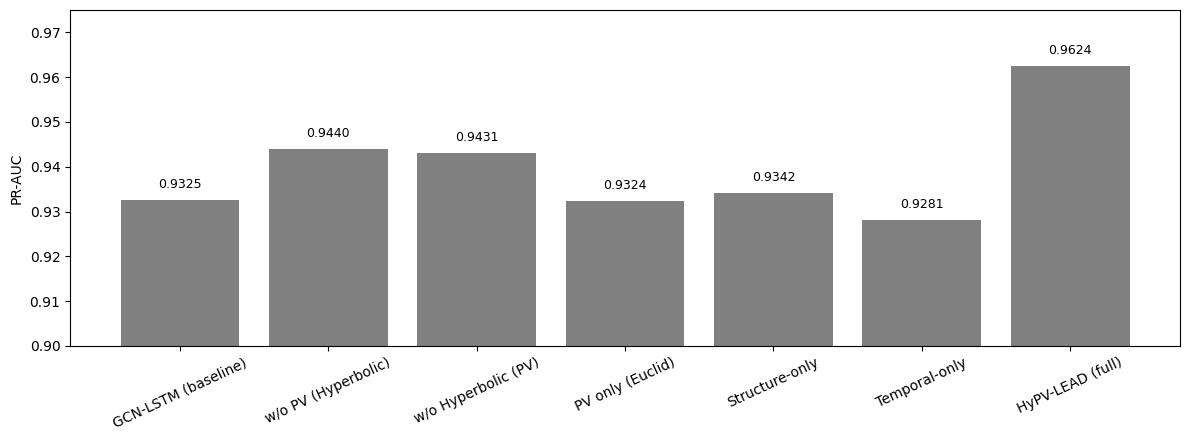}
    \caption{Ablation Study Results}
    \label{fig:placeholder}
\end{figure}

The detailed ablation results are summarized in Table~\ref{tab:ablation-results} and further illustrated in Fig.~\ref{fig:placeholder}, which provide an intuitive comparison across model variants.

\begin{table}[t]
\centering
\caption{Ablation Study Results}
\label{tab:ablation-results}
\resizebox{\columnwidth}{!}{%
\begin{tabular}{|l|c|c|}
\hline
\textbf{Model Setting} & \textbf{PR-AUC} & \textbf{$\Delta$ vs Full} \\
\hline
GCN-LSTM (Baseline)            & 0.9325 & -0.0299 \\ 
w/o PV (Hyp + GCN-LSTM)        & 0.9440 & -0.0184 \\ 
w/o Hyperbolic (PV + GCN-LSTM) & 0.9431 & -0.0193 \\ 
PV only (Euclid + GCN-LSTM)    & 0.9324 & -0.0300 \\ 
Structure-only (PV + Hyp + GCN)& 0.9342 & -0.0282 \\ 
Temporal-only (PV + Hyp + LSTM)& 0.9281 & -0.0343 \\ 
Full HyPV-LEAD                  & 0.9624 & +0.0000 \\ 
\hline
\end{tabular}}
\end{table}

\section{Conclusion}

The detection of abnormal cryptocurrency transactions has long been constrained by post hoc, model-oriented approaches that identify events only after they occur. While these methods have provided incremental technical refinements, they fall short in practice: rare anomalies are often missed, temporal dynamics are disrupted, and regulators or exchanges receive alerts too late to act. In such environments, where transaction volumes and manipulation tactics evolve rapidly, what is most needed is not another complex classifier but a framework that delivers actionable warnings in advance. 

\par This study addressed that gap through \textit{HyPV-LEAD}, a framework that redefines anomaly detection as a proactive early-warning task. By explicitly coupling observation windows with lead-time horizons, the framework ensures that signals of irregular behavior are detected before they fully unfold. In parallel, the introduction of PV sampling allows the system to handle severe class imbalance without distorting temporal continuity, while hyperbolic embeddings preserve the hierarchical and clustered nature of blockchain transaction networks. Together, these elements create a structural--temporal representation that is both richer and more robust than conventional Euclidean or sequential-only methods. 

\par Extensive evaluation on large-scale Bitcoin transaction data confirmed the effectiveness of this approach. \textit{HyPV-LEAD} consistently outperformed state-of-the-art baselines across accuracy, recall, and PR-AUC, demonstrating that reliable early-warning is not only theoretically possible but practically achievable. Ablation studies further showed that each design choice contributes meaningfully to the whole, and that only through their integration can high levels of predictive reliability be sustained in the face of dynamic and imbalanced data. These findings indicate that the proposed framework moves the field beyond incremental gains toward a new standard of proactive anomaly detection. 

\par Although the primary evaluation focused on mixing-related behaviors, the framework is readily extensible to other forms of manipulation such as phishing, layering, and pump-and-dump operations. This adaptability highlights its potential impact beyond academic benchmarks, providing a pathway toward real-time monitoring systems that support anti-money laundering, financial security, and regulatory compliance in rapidly evolving digital markets. In conclusion, \textit{HyPV-LEAD} offers not only a methodological contribution but also a practical foundation for transforming blockchain anomaly detection from reactive analysis to genuine early-warning systems.

\section*{Acknowledgment}

This work was supported by the National Research Foundation of Korea (NRF) grant funded by the Korean government under the project “Socio-Technological Solutions for Bridging the AI Divide: A Blockchain and Federated Learning-Based AI Training Data Platform” (NRF-2024S1A5C3A02043653).

\bibliographystyle{ieeetr}   
\bibliography{references}

\end{document}